\documentclass[conference]{IEEEtran}
\IEEEoverridecommandlockouts
\usepackage{cite}
\usepackage{amsmath,amssymb,amsfonts}
\usepackage{algorithmic}
\usepackage{graphicx}
\usepackage{textcomp}
\usepackage{xcolor}
\usepackage{booktabs}
\usepackage{multirow}
\usepackage{url}

\def\BibTeX{{\rm B\kern-.05em{\sc i\kern-.025em b}\kern-.08em
    T\kern-.1667em\lower.7ex\hbox{E}\kern-.125emX}}
\begin{document}

\title{The Efficiency Frontier: A Unified Framework for Cost–Performance Optimization in LLM Context Management} 

\author{

\IEEEauthorblockN{
\begin{tabular}{c c}
\begin{tabular}{c}
Binqi Shen\textsuperscript{$\dagger$*}\\
\textit{Department of Industrial Engineering}\\
\textit{and Management Sciences}\\
\textit{Northwestern University}\\
Evanston, USA\\
binqishen2021@u.northwestern.edu
\end{tabular}
&
\begin{tabular}{c}
Lier Jin\textsuperscript{$\dagger$}\\
\textit{Fuqua School of Business}\\ 
\textit{Duke University}\\
Durham, USA\\
lierjin@alumni.duke.edu
\end{tabular}
\end{tabular}
}

\vspace{1.5em}

\IEEEauthorblockN{
\begin{tabular}{c c c}
\begin{tabular}{c}
Hanyu Cai\\
\textit{Department of Industrial Engineering}\\
\textit{and Management Sciences}\\
\textit{Northwestern University}\\
Evanston, USA\\
hanyucai2022@u.northwestern.edu
\end{tabular}
&
\begin{tabular}{c}
Lan Hu\\
\textit{Department of Engineering}\\
\textit{Carnegie Mellon University}\\
Pittsburgh, USA\\
lanh@alumni.cmu.edu
\end{tabular}
&
\begin{tabular}{c}
Yuting Xin\\
\textit{Department of Information}\\
\textit{and Decision Sciences}\\
\textit{University of Minnesota}\\
Minneapolis, USA\\
yuting.xin@outlook.com
\end{tabular}
\end{tabular}
}

\vspace{1.5em}

\IEEEauthorblockA{
\vspace{0.8em}
\textsuperscript{$\dagger$}Equal contribution \quad
\textsuperscript{*}Corresponding author
}
}

\maketitle

\begin{abstract} Large language models (LLMs) increasingly rely on long-context processing, but expanding context windows introduces substantial computational and financial costs. Existing context reduction approaches, including retrieval and memory compression methods, are typically evaluated using performance and efficiency metrics independently, limiting systematic comparison and deployment-aware decision-making.

This paper introduces \textit{The Efficiency Frontier}, a unified framework for cost--performance optimization in LLM context management. The framework models context strategy selection as a deployment-aware optimization problem that jointly accounts for task performance, token cost, and preprocessing reuse through amortized cost modeling. Unlike existing evaluations that compare methods in isolation, the proposed framework enables decision-oriented analysis. It identifies when different context management strategies become preferable under varying operational conditions. Experiments on HotpotQA reveal distinct operational regimes and transition boundaries between retrieval-based and preprocessing-based strategies. Results show that deployment-aware optimization reduces effective token usage by approximately 25\% at comparable performance, enabling more cost-efficient deployment of large language model systems, while amortized memory compression achieves over 50\% lower token cost relative to full-context prompting in higher-performance settings. Overall, the proposed framework provides a principled and practical foundation for evaluating and deploying scalable, efficient, and sustainable LLM systems across enterprise, scientific, and public-sector applications.
\end{abstract}


\section{Introduction}

Large language models (LLMs) have achieved rapid progress in recent years, demonstrating strong capabilities across a wide range of natural language processing tasks, such as search, customer support, and knowledge work \cite{Raza2025wu}. However, these advances come with increasing computational and financial costs, driven by both model scale and the growing length of input context \cite{zhang2023dissectingruntime}. As context windows continue to expand, the computational cost of processing additional tokens often increases faster than the corresponding gains in downstream task performance, making efficient context utilization an increasingly important challenge \cite{2025wuchen-evaluation}. At the same time, the environmental impact of large-scale AI systems, including energy and water consumption, has raised growing concerns about their long-term sustainability \cite{wu2022sustainable}. These challenges highlight the need for more efficient use of context in LLM systems, particularly as large language models are increasingly deployed in enterprise, scientific, and public-sector applications where computational cost, scalability, and resource utilization directly affect operational feasibility.

Recent work has explored a variety of techniques for reducing context length while preserving task performance, including retrieval-based filtering, summarization, and context compression methods \cite{jiang2024longllmlingua,xu2024retrieval}. These approaches aim to improve efficiency by selectively retaining the most relevant information while discarding redundant or less informative content \cite{2026jiangmagma}. Although these methods have shown promising results, their evaluation remains fragmented. Existing studies typically report performance metrics such as exact match (EM) or F1 score, alongside cost indicators such as token usage or latency \cite{liang2023holisticevaluationlanguagemodels}. However, these metrics are often considered independently and rarely provide a unified assessment of the trade-off between cost reduction and performance degradation \cite {pollertlam2026context}. Moreover, retrieval, compression, and long-context approaches are frequently evaluated under different experimental settings, making direct comparison difficult. As a result, it remains difficult to systematically compare different context reduction strategies or assess when one strategy should be preferred over another under practical deployment constraints \cite{2026jiangdpongming}.

To address this limitation, we propose a unified evaluation framework for methodically assessing the efficiency of context reduction techniques in large language models. We introduce the concept of the \textbf{Efficiency Frontier}, a three-stage evaluation framework that characterizes the trade-off between task performance and computational cost across different context management strategies. Unlike existing approaches that evaluate performance and cost in isolation, our framework provides explicit decision logic for selecting context management strategies, bridging the gap between retrieval-based methods and long-context processing. The framework incorporates a parameterized log-utility metric to model diminishing returns from additional context while accounting for amortized preprocessing cost. By varying a reuse parameter ($N$), the framework supports systematic comparison under realistic deployment constraints by identifying crossover regions where different strategies become preferable.

Beyond evaluation, the framework provides practical guidance for context management strategy selection under varying cost and reuse conditions. We validate the framework on HotpotQA \cite{yang2018hotpotqa}, a multi-hop question answering benchmark containing both relevant and distractor context. More broadly, the framework supports deployment-aware optimization by enabling practitioners to balance task quality and computational resource requirements under real-world operating constraints.

\section{Related Work}

\subsection{Evaluation of Large Language Models}

Recent evaluation frameworks for large language models have expanded beyond task accuracy to incorporate additional dimensions such as robustness, fairness, computational efficiency, and sensitivity to prompting conditions and interaction style. Beyond task performance, frameworks like HELM and recent specialized benchmarks increasingly emphasize multi-dimensional evaluation of model behavior, particularly regarding the trade-offs between accuracy and execution efficiency \cite{liang2023holisticevaluationlanguagemodels, 2023zhaobenchmark}. At the same time, work on efficient and sustainable AI has highlighted the importance of resource-aware evaluation criteria, including computational cost, energy consumption, and latency \cite{2025raodatacentric}. For example, Green AI advocates for incorporating efficiency and resource usage into model evaluation as model scale and deployment costs continue to increase \cite{schwartz2020green}. Beyond general behavior and resource usage, recent work has identified the need for evaluating the suitability of alignment systems, defined as their reliability under real-world perturbations \cite{2025zangreward}. This shift underscores the necessity of moving beyond static benchmarks toward evaluation frameworks that are verifiably robust under deployment conditions. 

However, existing methodologies typically treat task effectiveness, computational cost, and deployment efficiency as independent variables. This fragmentation obscures the practical trade-offs involved in real-world deployment, where practitioners must balance task performance and computational cost without standardized evaluation criteria \cite{2025sunwenxihighrecalldeeplearning, cao2026taskspecificefficiencyanalysissmall}. Many studies report performance metrics such as F1 or compression ratio alongside basic cost indicators, but rarely provide deployment-aware, end-to-end comparisons of per-query token or monetary cost against task performance across different context management strategies. This limitation becomes particularly important in long-context settings, where increases in context length can substantially increase computational cost without consistent gains in downstream performance. Recent work on long-context evaluation shows that increasing context or model complexity does not necessarily yield proportional performance gains \cite{liu2024lost}.

\subsection{Context Length Scaling and Diminishing Returns}

As long-context capabilities continue to expand, recent advances in large language models (LLMs) have significantly increased maximum context length, enabling models to process longer sequences and incorporate more information into their reasoning process. While longer context windows can improve performance in tasks requiring multi-hop reasoning or long-range dependencies, empirical evidence suggests that these gains are often subject to diminishing returns \cite{du2025contextlengthhurtsllm}.

Research has shown that LLMs do not always effectively utilize long input sequences. The “lost in the middle” phenomenon demonstrates that models tend to underutilize information located in the middle of long sequences \cite{liu2024lost}, while more recent studies report degraded performance due to attention dilution and distractor interference as context length increases \cite{bansal2025let, gu2026long}. Large-scale evaluations further show that models often fail to fully utilize the additional context available to them \cite{chen2026longbench}.

At the same time, the computational cost of long-context processing grows disproportionately with sequence length due to the quadratic complexity of attention mechanisms \cite{vaswani2017attention}, whereas performance improvements are often sublinear or inconsistent \cite{liu2024lost, bai2024longbench,2026laitransformers}. These limitations have motivated growing interest in methods that reduce or selectively process context in order to improve efficiency while preserving task performance.
Existing work, however, primarily focuses on improving long-context capabilities or benchmarking performance, rather than on systematically modeling the trade-offs among context length, computational cost, and downstream performance.

\subsection{Context Reduction Techniques}

To mitigate the high computational cost associated with long-context processing, a growing body of work has explored techniques for reducing context length while preserving task performance. Recent work has proposed various context compression techniques, including token-level compression strategies and instruction-driven routing mechanisms that selectively sparsify input tokens to reduce inference latency \cite{ge2023context, 2025licogvla}. Other studies explore reasoning-enhanced adaptation, instruction tuning, and multimodal fusion strategies for improving context understanding and efficient context utilization in complex LLM settings \cite{2025wangzixuanreasoning, 2025sunyuaudioenhanced, li2025human}. Such strategies are increasingly employed to enable real-time deployment for time-constrained applications \cite{2024zhaobalf}. In addition, context reduction methods such as semantic sparsification and filtering techniques aim to remove redundant context prior to generation, improving efficiency, robustness, and risk-aware resilience \cite{2025lisemanticvla, 2025raodynamicsamplingadaptsiterative, 2026xueresilient}. Building on these ideas, hybrid retrieval and routing approaches have also been proposed to further improve robustness and context selection \cite{2026chengresolvingrobustness}.

Existing work primarily evaluates retrieval, compression, and long-context processing in isolation, with comparisons often performed under different datasets, prompting settings, or cost assumptions. As a result, it remains difficult to determine when one strategy is more efficient or effective than another under comparable conditions. This lack of standardized evaluation makes it challenging to reason systematically about efficiency-performance trade-offs across context management strategies, motivating the need for a unified evaluation framework.

\section{Methodology}

We propose a structured, three-stage framework for systematically evaluating the trade-off between performance and computational cost in context management strategies for large language models. Unlike prior approaches that optimize accuracy or efficiency in isolation, our framework explicitly models \textit{decision-making under deployment constraints}, enabling strategy selection conditioned on both performance requirements and system usage patterns.

A central contribution is the distinction between \textit{intrinsic cost} (per-query inference cost) and \textit{amortized cost} (including reusable preprocessing), captured through a reuse parameter $N$. This formulation reflects realistic deployment settings, such as shared memory systems, cached summaries, and multi-query workloads, where upfront computation can be reused across queries. As a result, the framework supports evaluation across heterogeneous operational regimes within a unified objective.

\subsection{Efficiency Frontier Framework}

We formulate context management as a \textit{decision problem}: given a deployment preference over performance and cost, select the strategy and configuration that maximizes utility.

\subsubsection{\textbf{Cost Model}} We model computation as a two-stage process. Let $T_{\text{stage1}}$ denote context preprocessing cost (e.g., memory compression), and $T_{\text{stage2}}$ denote per-query inference cost. When context preprocessing is reused across $N$ queries, the effective cost is:
\begin{equation}
\text{EffectiveTokens} = T_{\text{stage2}} + \frac{T_{\text{stage1}}}{N}
\label{eq:equation1}
\end{equation}

This distinguishes \textit{intrinsic cost} (per-query inference) from \textit{amortized cost} under context reuse.

\subsubsection{\textbf{Efficiency Score}}
We define a parameterized utility function that captures the trade-off between performance and cost:
\begin{equation}
\text{EfficiencyScore}(w) = w \cdot F1 - (1 - w) \cdot \log(\text{EffectiveTokens})
\label{eq:equation2}
\end{equation}
where $w \in [0,1]$ controls the preference between performance and efficiency. Larger $w$ emphasizes accuracy, while smaller $w$ prioritizes lower cost. In practice, $w$ reflects deployment-specific priorities rather than a fixed universal setting. Cost-sensitive applications may favor smaller values of $w$, whereas quality-critical applications may favor larger values. Rather than prescribing a single optimal value, the proposed framework evaluates a range of $w$ values to identify strategy transition points and operating regions that align with deployment requirements.

This formulation captures two key properties: (i) amortization of preprocessing cost under reuse via $N$, and (ii) diminishing sensitivity to token cost through the logarithmic penalty, reflecting practical tolerance to cost increases at scale.

\subsubsection{\textbf{Optimization Procedure}} The Efficiency Frontier is constructed in three stages:

\begin{itemize}
    \item Stage 1: Intra-Strategy Optimization.
For each strategy, we evaluate a range of configurations (e.g., compression ratios, retrieval depth) and retain only those that are optimal under some $w$:

\begin{equation}
\arg\max_{\text{config}} \; \text{EfficiencyScore}(w)
\end{equation}
This step removes dominated configurations while preserving configurations that may be optimal under at least one preference setting.

    \item Stage 2: Candidate Scoring and Evaluation. 
All retained configurations are evaluated under the amortized cost formulation above, ensuring consistent comparison across strategies with heterogeneous cost structures.

    \item Stage 3: Global Decision Optimization.
We aggregate candidates across all strategies and compute the globally optimal choice:
\begin{equation}
\arg\max_{\text{strategy, config}} \; \text{EfficiencyScore}(w)
\end{equation}

Sweeping over $w$ yields a sequence of decision transition points, forming the global Efficiency Frontier. This frontier induces (i) a strategy transition map across preference regimes, and (ii) a lookup table mapping target performance levels to minimum achievable cost. 

\end{itemize}

\subsection{Role of Amortization ($N$)}

The reuse parameter $N$ models repeated use of preprocessing outputs across queries. As $N$ increases, preprocessing costs are amortized, making higher-cost strategies such as memory compression increasingly competitive. This enables the framework to capture deployment scenarios ranging from single-turn interactions to persistent memory systems.

\subsection{Context Management Strategies}

We evaluate representative strategies that span distinct mechanisms of context utilization and cost structures:

\begin{itemize}
    \item Full-Context Prompting: concatenation of all available context. This serves as a high-cost baseline that maximizes information availability and approximates upper-bound performance under unconstrained context.
    
    \item Oracle Retrieval: construction of context using only ground-truth supporting documents. This provides a non-deployable upper bound that isolates limitations due to context selection from those due to model reasoning.
    
    \item Memory Compression: LLM-based preprocessing that transforms raw context into a condensed representation prior to inference. This introduces explicit Stage 1 cost while reducing Stage 2 cost, enabling analysis of trade-offs between compression fidelity and computational overhead.
    
    \item Zero-Cost Retrieval: TF-IDF (vanilla and query-aware) and semantic embedding retrieval. Vanilla TF-IDF ranks sentences based solely on corpus statistics, independent of the input question, while query-aware TF-IDF incorporates the input question to prioritize context relevant to the specific task. Semantic embedding retrieval encodes both the Full-Context and question into dense vector representations and retrieves the top-$k$ documents based on semantic similarity. These methods perform context selection without LLM-based preprocessing, incurring no Stage 1 token cost and establishing a lower bound on computational overhead.
\end{itemize}

Together, these strategies cover a broad design space, including full-context baselines, upper-bound references, model-based preprocessing, and lightweight retrieval approaches. This diversity enables thorough analysis of how different mechanisms interact with cost, performance, and reuse. The present study focuses on deployment-level context management techniques that can be applied without modifying the underlying model architecture. Architectural approaches, including dynamic sparse attention and other efficient long-context architectures, represent a complementary direction for reducing context-processing cost. Evaluating such methods within the proposed framework remains an important direction for future work.

\subsection{Evaluation Setup}

We instantiate the framework on HotpotQA, a multi-hop question answering benchmark containing both relevant and distractor context, making it a representative testbed for evaluating context management strategies.

We evaluate 5,000 randomly sampled HotpotQA instances (seed = 42), and report 95\% bootstrap confidence intervals over question-level F1 scores. This sample size was selected to provide stable aggregate performance estimates while remaining computationally feasible given the large number of strategy and configuration combinations evaluated. 

All strategies are evaluated using GPT-5.4 mini (OpenAI API, accessed May 2026) under a standardized prompt and deterministic inference configuration to isolate the effect of context management. GPT-5.4 mini was selected to balance reasoning capability and computational scalability: the model is sufficiently strong to support reliable multi-hop reasoning while remaining efficient enough to enable large-scale evaluation across thousands of strategy and configuration combinations.

\section{Experiments}

We apply the proposed framework to analyze how context management strategies behave under varying performance requirements and deployment constraints. Rather than comparing strategies in isolation, the framework reveals how optimal choices depend jointly on target performance, computational budget, and reuse patterns.

Results are presented through three complementary analyses: (i) Efficiency Frontier Analysis, which characterizes intra- and inter-strategy trade-offs, (ii) Decision Patterns Across Regimes, which maps performance targets to deployment-dependent optimal strategies, and (iii) Learnings and Practical Implications, which distills broader system-level insights for LLM deployment. Unless otherwise noted, reported F1 values represent mean performance over the 5,000-instance evaluation set. Uncertainty estimates are reported as 95\% bootstrap confidence intervals. All reported token costs correspond to effective per-question token usage after amortizing preprocessing costs according to Eq.~\eqref{eq:equation1} and averaged over the evaluated HotpotQA instances.

\subsection{Efficiency Frontier Analysis}

\begin{figure*}[t]
    \centering
    \includegraphics[width=0.9\textwidth]{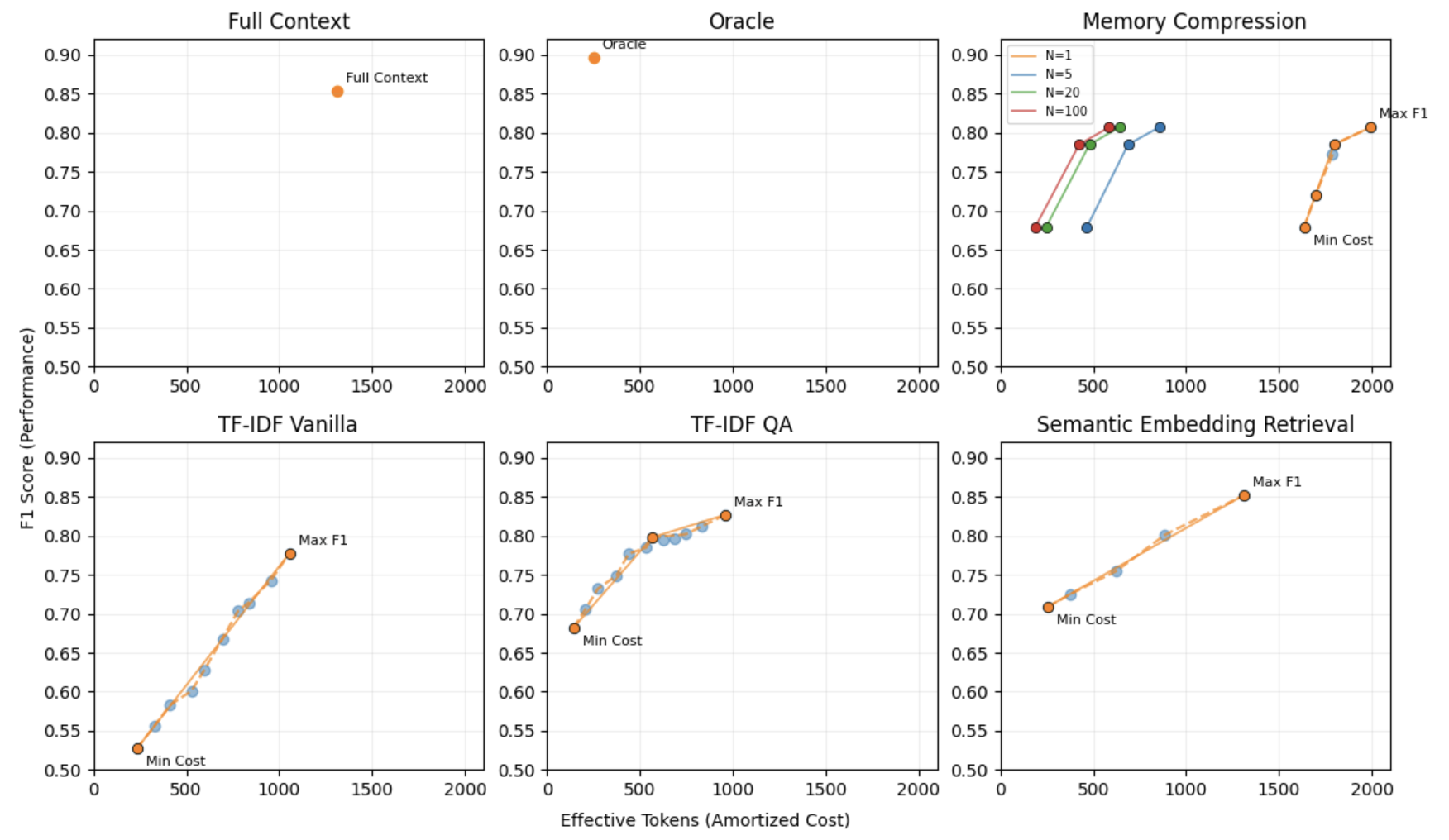} 
     \caption{Strategy-level Efficiency Frontiers and decision paths. Each panel plots token cost versus task performance (F1). Faint points denote all evaluated configurations, dashed lines indicate the Pareto frontier, and solid lines trace optimal configurations under varying preference weight $w$. Across strategies, the optimal operating point changes continuously with preference weight $w$, indicating that no single configuration is universally optimal across deployment settings.}
    \label{fig:stage12}
\end{figure*}

\begin{figure*}[t]
    \centering
    \includegraphics[width=0.9\textwidth]{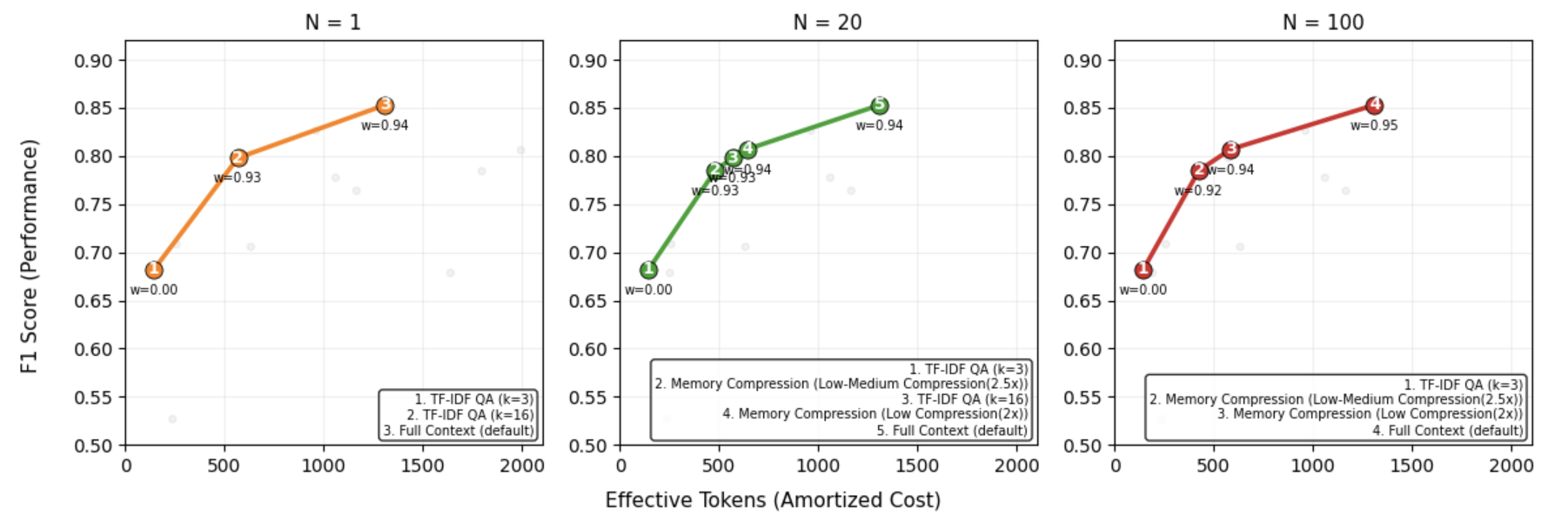} 
    \caption{Global Efficiency Frontier under different reuse regimes ($N$). Each curve traces the sequence of optimal (strategy, configuration) choices as preference weight $w$ varies. As reuse increases, preprocessing-based methods such as memory compression become optimal across a broader range of preference weights, replacing lightweight retrieval methods in parts of the frontier.}
    \label{fig:stage3}
\end{figure*}

{Fig.~\ref{fig:stage12}} provides a unified view of strategy-level Efficiency Frontiers and decision paths. Each panel captures three layers: (i) the complete configuration space (all evaluated parameter settings), (ii) the intrinsic Pareto frontier (non-dominated trade-offs), and (iii) the optimal operating path induced by the efficiency score.

This reveals a key structural property: \textit{each strategy admits a single intrinsic frontier, but multiple deployment-dependent operating points}. As preference weight $w$ shifts, the optimal configuration moves along the frontier rather than remaining fixed.

\begin{itemize}

\item Query-aware retrieval improves the intrinsic frontier.
Compared to vanilla TF-IDF, query-aware retrieval consistently achieves higher performance at comparable cost, shifting the Pareto frontier upward without increasing preprocessing overhead.

\item Memory compression becomes increasingly competitive under amortization.
Unlike retrieval methods, memory compression introduces substantial preprocessing cost. However, as reuse increases, amortization reduces its effective cost, causing memory compression to occupy a larger portion of the global frontier. 

\end{itemize}

{Fig.~\ref{fig:stage3}} aggregates these frontiers into a global view. Each curve traces the sequence of optimal (strategy, configuration) choices as preference weight $w$ shifts from cost-sensitive to performance-sensitive regimes, with transition points indicating where one strategy becomes more efficient than another.

A consistent pattern emerges: higher reuse ($N$) favors strategies with higher upfront cost. As $N$ increases, amortization reduces effective cost, enabling memory compression to dominate across broader balanced regimes. This demonstrates that optimal strategy selection is inherently deployment-dependent rather than fixed: isolated single-query workloads tend to favor lightweight retrieval methods, whereas persistent assistants and enterprise knowledge systems increasingly benefit from preprocessing-heavy approaches such as memory compression.

Bootstrap confidence intervals for representative frontier configurations are reported in Table~\ref{tab:ci}. The intervals remain narrow despite the heterogeneous difficulty of HotpotQA examples, indicating that the observed frontier structure and strategy transition patterns are stable.

\begin{table}[htbp]
\caption{Bootstrap confidence intervals for representative frontier configurations.}
\label{tab:ci}
\centering
\footnotesize
\begin{tabular}{lcc}
\hline
Configuration & Mean F1 & 95\% CI \\
\hline
TF-IDF QA ($k=16$) & 0.78 & [0.771, 0.789] \\
Memory Compression ($2.5\times$) & 0.78 & [0.772, 0.790] \\
Memory Compression ($2\times$) & 0.80 & [0.792, 0.808] \\
Full-Context & 0.82 & [0.811, 0.829] \\
\hline
\end{tabular}
\end{table}

\subsection{Decision Patterns Across Regimes}

While the Efficiency Frontier provides a continuous view of optimal decisions across preference weights, practitioners often require discrete guidance: given a target performance level, which strategy should be selected? To address this, we translate the frontier into a decision-oriented view, mapping performance targets to optimal strategies under different deployment conditions.

To better understand how the framework informs decision-making, we group operating points into three practical regimes based on performance requirements: (i) efficiency-oriented regime, (ii) balanced regime, and (iii) high-performance regime. Table~\ref{tab:lookup} summarizes representative operating points from the global frontier, translating the continuous frontier into a discrete, deployment-oriented decision guide.

\begin{itemize}

\item \textbf{Efficiency-oriented regime ($F1 < 0.78$).}
Lightweight retrieval methods dominate due to minimal cost. In low-reuse settings, TF-IDF QA achieves competitive performance at the lowest token usage.

\item \textbf{Balanced regime ($0.78 \leq F1 < 0.82$).}
This regime exhibits the most variation across deployment settings. Under higher reuse, memory compression becomes increasingly favorable, often becoming competitive or dominant in balanced regimes.

\item \textbf{High-performance regime ($F1 \geq 0.82$).}
Full-Context prompting remains necessary to achieve peak performance. However, this comes at a significant cost increase, often exceeding 2$\times$ that of balanced configurations, indicating diminishing returns.

\end{itemize}

\begin{table}[htbp]
\caption{Dominant strategy across performance regimes and reuse levels.}
\label{tab:lookup}
\centering
\footnotesize
\setlength{\tabcolsep}{2.5pt}
\renewcommand{\arraystretch}{0.95}

\begin{tabular}{c c c}
\hline
\textbf{Regime (F1 Range)} & \textbf{$N=1$} & \textbf{$N=100$} \\
\hline

\shortstack[c]{Efficiency-oriented\\($0.70$--$0.78$)}
& TF-IDF QA ($k{=}16$)
& Mem. Comp. ($2.5\times$) \\

\shortstack[c]{Balanced\\($0.78$--$0.82$)}
& Full-Context
& Mem. Comp. ($2\times$) \\

\shortstack[c]{High-performance\\($0.82$--$0.84$)}
& Full-Context
& Full-Context \\
\hline
\end{tabular}
\end{table}

The framework enables direct quantification of efficiency gains across regimes:

\begin{itemize}
    \item In the Balanced regime, increasing reuse from $N=1$ to $N=100$ shifts the optimal operating point from TF-IDF QA (566 EffectiveTokens) to memory compression (424 EffectiveTokens), reducing effective cost by approximately 25\% at comparable performance ($F1 = 0.78$), with overlapping confidence intervals suggesting comparable task performance. 
    \item In the High-performance regime ($F1 \geq 0.82$), full-context prompting remains necessary, but incurs substantial cost: achieving the highest evaluated performance levels with full-context prompting requires more than 2$\times$ the EffectiveTokens cost of balanced-regime operating points. This highlights a consistent pattern of diminishing returns at higher performance levels.
\end{itemize}

While the specific transition points and thresholds reported here are derived from HotpotQA, the observed structure, namely the existence of distinct efficiency regimes and strategy transition boundaries, arises from the underlying cost–performance trade-off and is expected to generalize across tasks with similar context characteristics.

\subsection{Learnings and Practical Implications}

\subsubsection{\textbf{A unified framework is necessary}}
Analysis across all configurations reveals a strongly non-linear relationship between performance and token cost. Achieving higher performance requires disproportionately larger increases in computation: for example, improving from mid-range performance ($F1 \approx 0.78$) to high performance ($F1 \approx 0.84$) more than doubles token usage. 

Evaluating strategies using performance or cost in isolation therefore obscures deployment-dependent trade-offs and can lead to misleading conclusions. The proposed efficiency metric provides a unified view, enabling principled comparison and decision-making across competing strategies.

\subsubsection{\textbf{System-level efficiency gains}}
Aligning strategy selection with deployment conditions yields substantial efficiency improvements. In our setting, this translates to approximately 25\% reduction in token usage at comparable performance in the balanced regime ($F1 \approx 0.78$) when moving from a low-reuse setting ($N=1$) to a high-reuse setting ($N=100$), shifting the optimal operating point from TF-IDF QA (566 EffectiveTokens) to Memory Compression (424 EffectiveTokens). Similarly, near the upper end of the balanced regime ($F1 \approx 0.80$), increasing reuse from $N=1$ to $N=100$ shifts the optimal strategy from Full-Context (1308 EffectiveTokens) to Memory Compression (584 EffectiveTokens), yielding over 50\% effective cost reduction. These gains arise not from improving individual strategies, but from selecting the right strategy under the right conditions.

\subsubsection{\textbf{Strategy selection is deployment-dependent}}
No single strategy dominates across all scenarios. Lightweight retrieval methods are optimal in low-cost regimes, while preprocessing-based approaches such as memory compression become increasingly favorable when reuse is present. At the highest performance levels, Full-Context remains necessary despite its cost. This highlights the importance of incorporating deployment factors, such as reuse and performance targets, into evaluation.

\subsubsection{\textbf{Operational guidance}}
The framework supports two complementary modes of use. First, sweeping the preference parameter $w$ provides a continuous view of trade-offs between performance and cost. Second, the decision table offers a discrete mapping from target performance to optimal strategy, enabling straightforward integration into system design and deployment pipelines. This capability is particularly valuable for organizations deploying LLM-based applications at scale, where resource constraints, inference costs, and service quality requirements must be balanced simultaneously.

\subsubsection{\textbf{Implications for sustainable AI systems}}
By enabling systematic reductions in unnecessary computation, the framework provides a practical pathway toward more efficient and sustainable deployment of large-scale language models. Rather than scaling context indiscriminately, it supports targeted context utilization, reducing computational overhead while preserving task performance.

\section{Conclusion}

This work proposes the \textbf{Efficiency Frontier}, a unified framework for evaluating context management strategies in large language models under explicit performance--cost trade-offs. Rather than treating accuracy and computational efficiency as separate objectives, the framework models strategy selection as a deployment-dependent decision problem, jointly accounting for task performance, inference cost, and amortized preprocessing reuse.

Across experiments on HotpotQA, several consistent patterns emerge. First, the relationship between performance and computational cost is strongly non-linear: achieving incremental gains in performance often requires disproportionately larger increases in token usage. Second, no single context management strategy is universally optimal. Lightweight retrieval methods dominate efficiency-oriented regimes, while preprocessing-based methods such as memory compression become increasingly favorable under high reuse settings due to amortization effects. Finally, full-context prompting remains necessary for peak performance, but exhibits clear diminishing returns relative to its computational cost.

These findings address several limitations in existing evaluations of context reduction methods. The proposed framework provides (i) a unified objective that jointly models performance and cost, (ii) explicit operational transition points between competing strategies, and (iii) deployment-aware recommendations that are conditioned on reuse patterns and performance targets. By incorporating amortized preprocessing cost through the reuse parameter $N$, the framework also captures practical deployment scenarios that are often overlooked in existing benchmarking settings, including persistent memory systems, shared retrieval pipelines, and multi-query inference workloads.

Beyond evaluation, the framework provides practical value for both academic research and real-world deployment. By enabling systematic reductions in computational requirements while preserving task quality, the proposed framework supports the development of more scalable and cost-efficient AI systems suitable for real-world deployment across a wide range of application domains. In research settings, the Efficiency Frontier offers a standardized methodology for comparing heterogeneous context management approaches under a common decision framework, enabling more reproducible and deployment-relevant evaluation of emerging LLM optimization methods. In industry settings, the framework can support system-level optimization by guiding strategy selection according to operational constraints such as latency budgets, token cost limits, and query reuse patterns. Experimental results demonstrate that deployment-aware strategy selection can reduce EffectiveTokens usage by approximately 25\% at comparable performance in mid-range operating regimes, and by more than 50\% relative to full-context baselines under amortized reuse conditions. These improvements are achieved not through larger models or additional training, but through more efficient utilization of context.
 
More broadly, the proposed framework contributes toward the development of more sustainable large-scale AI systems. As LLM deployment continues to scale across enterprise, scientific, and public-sector applications, computational efficiency is becoming increasingly important not only for economic reasons, but also for infrastructure scalability and environmental sustainability. By formalizing the trade-off between performance and context cost, the Efficiency Frontier shifts the focus from maximizing context length toward optimizing context utilization, providing a practical foundation for more efficient and scalable AI infrastructure.

Future work may extend the framework to additional long-context tasks, incorporate system-level objectives such as latency, energy consumption, and monetary cost, and explore adaptive utility functions that better capture application-specific deployment priorities. Future research may also investigate more specialized context-management approaches, including methods that leverage structured domain knowledge and domain-aware representation learning. Prior work has shown that tailored latent modeling frameworks and optimization objectives can improve representation fidelity and capture complex underlying relationships in high-dimensional systems \cite{2023yansemantic, 2025yannew}.

Overall, the proposed Efficiency Frontier framework establishes a principled and practical foundation for deployment-aware optimization of context utilization in large language model systems.

\bibliographystyle{IEEEtran} 
\bibliography{main}    

\end{document}